%% file: main.tex
\documentclass[11pt]{article}

\usepackage[preprint]{acl}
\newif\ifdiffanon \diffanonfalse
\usepackage{times}
\usepackage{latexsym}
\usepackage[utf8]{inputenc}
\usepackage[T1]{fontenc}
\usepackage{url}
\usepackage{booktabs}
\usepackage{amsfonts}
\usepackage{amsmath}
\usepackage{nicefrac}
\usepackage{microtype}
\usepackage{graphicx}
\usepackage{subcaption}
\usepackage{xcolor}
\usepackage{listings}
\usepackage{pgfplots}
\pgfplotsset{compat=1.18}
\usepackage{tikz}
\usetikzlibrary{decorations.pathreplacing,calc,arrows.meta}
\usepackage{colortbl}
\usepackage{array}
\usepackage{multirow}
\usepackage{makecell}
\usepackage{placeins}

\renewcommand{\arraystretch}{0.98}
\definecolor{kwcolor}{HTML}{005F87}
\definecolor{strcolor}{HTML}{6A7F00}
\definecolor{cmtcolor}{HTML}{5C5C5C}
\lstdefinestyle{mystyle}{
    basicstyle=\ttfamily\footnotesize\color{black},
    keywordstyle=\color{kwcolor}\bfseries,
    stringstyle=\color{strcolor},
    commentstyle=\color{black!62}\itshape,
    identifierstyle=\color{black},
    columns=fullflexible,
    breaklines=true,
    breakatwhitespace=false,
    captionpos=b,
    keepspaces=true,
    showspaces=false,
    showstringspaces=false,
    showtabs=false,
    tabsize=2,
    frame=single,
    rulecolor=\color{black!28},
    framesep=3pt,
    aboveskip=4pt,
    belowskip=4pt,
}
\lstdefinestyle{promptstyle}{
    basicstyle=\ttfamily\footnotesize\color{black},
    columns=fullflexible,
    breaklines=true,
    breakatwhitespace=false,
    keepspaces=true,
    showspaces=false,
    showstringspaces=false,
    showtabs=false,
    tabsize=2,
    frame=none,
    aboveskip=4pt,
    belowskip=4pt,
}
\newcolumntype{L}[1]{>{\raggedright\arraybackslash}p{#1}}

\newcommand{\llada}{LLaDA-8B}
\newcommand{\dream}{Dream-7B}
\newcommand{\masktoken}{\texttt{[MASK]}}

\newcommand{\repourl}{\href{https://github.com/lucky-verma/top1-fails-dlm-lora-monitors}{GitHub repository}}
\newcommand{\dataurl}{\href{https://github.com/lucky-verma/top1-fails-dlm-lora-monitors/tree/main/results}{result artifacts}}
\newcommand{\hpccluster}{CHIP HPC (UMBC)}
\newcommand{\hworkstation}{an H100 NVL (96GB) workstation}

\title{When Top-1 Fails:\\ Calibrating LoRA Monitors for Masked Diffusion LMs}

\author{
  Lucky Verma \\
  Independent Researcher \\
  \texttt{luckyv1@umbc.edu}
  \And
  Pratik Yadav \\
  University of Maryland, Baltimore County \\
  \texttt{pratiky1@umbc.edu}
}

\begin{document}

\maketitle

\begin{abstract}
Discrete diffusion language model (DLM) fine-tuning inherits inexpensive diagnostics from denoising-time confidence monitors, but their PEFT-training meaning is untested. We test top-$1$ argmax concentration as a collapse warning. Across $816$ LoRA/PEFT configurations from three DLM families, the warning fires for every configuration while logs record $0/816$ actual collapses at the $200$-step horizon, giving \textbf{zero precision}. The cause is pre-equilibrium saturation: top-$1$ concentration is already high before optimization and quickly becomes insensitive to final training stability. We then evaluate max LoRA gradient norm, a parameter-side signal that samples gradient routing rather than token concentration. On a pooled held-out LLaDA-family split, a train-optimized threshold identifies top-decile final-loss configurations with precision $0.68$ and $F_1{=}0.79$, above the all-positive top-$1$ baseline even at the lower split-bootstrap confidence bound. Autoregressive controls and cross-family threshold failures bound the result to short-horizon DLM-LoRA inspection rather than a universal collapse detector. Workflow: drop top-$1$ as a PEFT alarm, log max-gradient early in training, and calibrate thresholds per DLM family before routing runs for inspection.
\end{abstract}

\begin{figure*}[t]
\centering
\includegraphics[width=0.99\textwidth]{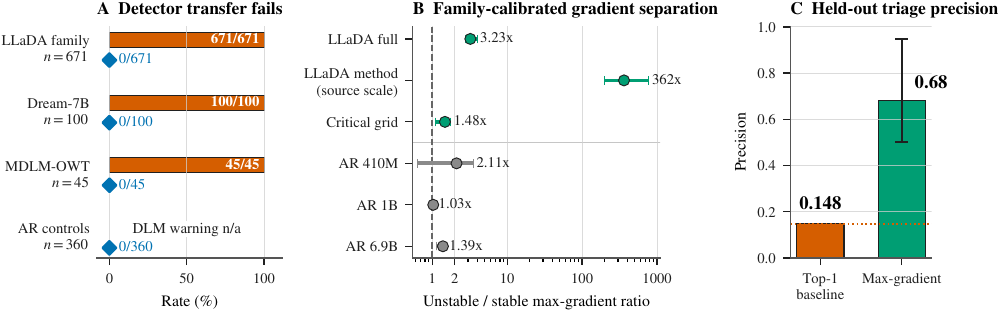}
\caption{\textbf{The transferred top-$1$ warning has zero precision, while max-gradient gives a LLaDA-family triage signal.}
(A) Across the $816$ DLM PEFT configurations, the top-$1$ warning fires in every configuration and observed collapse is $0/816$; AR controls have $0/360$ collapses and no top-$1$ warning by definition. (B) Stable-vs-unstable max-gradient effect sizes are large in the LLaDA-family DLM cohorts ($3.23\times$, $362\times$ on the source-scale method-comparison set, and $1.48\times$; Mann--Whitney $U$ with Bonferroni $m{=}6$ and bootstrap CIs), while AR controls are smaller or non-portable. (C) On a fixed held-out LLaDA-family split ($n{=}671$), max-gradient precision is $0.68$ with split-bootstrap $95\%$ CI $[0.500, 0.947]$, compared with the all-positive top-$1$ baseline ceiling $0.148$ (recall $0.94$, $F_1{=}0.79$).}
\label{fig:hero-top1-maxgrad}
\end{figure*}

\section{Introduction}
\label{sec:intro}

Discrete diffusion language models (DLMs)~\citep{llada2025, mdlm2024, dream2025} reconstruct fully masked sequences through iterative denoising, using bidirectional context rather than left-to-right prediction. As DLM checkpoints and fine-tuning recipes spread~\citep{zhang2024scaling, quant_dllm_2026, fast_dllm_2026, lad2025, dare2026}, practitioners need low-cost monitors for short-run LoRA training. A tempting candidate is already exposed by the audited DLM runners: the \textbf{top-1 collapse rate}, which measures whether argmax predictions concentrate on a small token vocabulary. This signal is logged for denoising/remasking diagnostics, but its meaning under short-horizon PEFT is unclear. We test the transfer directly: can top-$1$ collapse serve as a PEFT stability warning, and if not, what family-local monitor is more useful for inspection?

\textbf{The transfer fails.} Across three DLM model families spanning $816$ DLM PEFT configurations (LLaDA-family, four cohorts, $n{=}671$ + Dream-7B boundary cohort $n{=}100$ + MDLM-OWT 130M boundary cohort $n{=}45$), the warning fires in $\mathbf{816/816}$ ($\mathbf{100\%}$) configurations, while actual training collapse, logged by the same training loop's \texttt{collapsed} flag, occurs in $\mathbf{0/816}$ ($\mathbf{0\%}$) at the $200$-step horizon. The diagnostic has \textbf{zero precision}. Matched AR controls on Pythia $\{410\text{M},1\text{B},2.8\text{B},6.9\text{B}\}$ and Qwen3.5-9B ($360$ audited configurations; App.~\ref{app:ar-control}) also show $0/360$ actual collapses, so the result does not indicate a generic masked-CE collapse phenomenon. The warning fails to transfer into the tested DLM-LoRA PEFT setting.

\textbf{The failure has a measured explanation.} Across the same $671$ configurations, top-$1$ token frequency is $\mathbf{0.83 \pm 0.13}$ at training step $0$; every configuration is already above $0.5$, the median configuration crosses $0.95$ within $\mathbf{4}$ optimizer steps, and the legacy fire-step is stability-agnostic (Mann--Whitney $U$: $p{=}0.20$, n.s.; Fig.~\ref{fig:preequilibrium-top1}). A parameter-side check at the worst rank-amplification corner gives the complementary measurement: per-token CE gradients are only modestly concentrated (Gini $0.29$, largest evaluated token-position CE-gradient share $1.5\%$), while LoRA-parameter gradients are concentrated (Gini $0.46$, one matrix carries $63.0\%$ of gradient mass; App.~\ref{app:parameter-routing}). Top-$1$ tracks token-side pre-equilibrium concentration; max gradient norm samples the parameter-side routing that separates stable from unstable runs.

We evaluate \textbf{max gradient norm} as a family-local triage signal with Mann--Whitney $U$ tests and Bonferroni correction across six analyzable families ($m{=}6$). On LLaDA2.0-mini ($n{=}144$), unstable configurations have $3.23\times$ higher median max-gradient norm than stable configurations ($p_{\text{Bonf}}{=}2.7\times10^{-7}$, bootstrap $95\%$ CI $[2.76, 3.97]$); on the method-comparison set ($n{=}395$), the ratio is $362\times$ (stable median $99.3$ vs.\ unstable median $35{,}960.4$ in the source scale; $p_{\text{Bonf}}{=}5\times10^{-21}$, CI $[202, 779]$). The key check is held-out performance. On a fixed $80/20$ split of the $671$-configuration LLaDA-family corpus, a threshold selected on training configurations predicts top-decile final-loss configurations on held-out configurations with precision $\mathbf{0.68}$, recall $\mathbf{0.94}$, and $F_1{=}0.79$, versus $0.13$ precision for the all-positive top-$1$ baseline on this fixed split. A separate split-bootstrap gives $95\%$ CI $[0.500, 0.947]$, disjoint from the split-bootstrap baseline ceiling $0.148$; each bootstrap replicate resamples configurations, redraws the train/test split, and reselects the threshold on train; even the lower CI bound exceeds $3\times$ the baseline, and the supported use is inspection and routing rather than a high-precision gate (Limitations). Separately, a $B{=}200$ random-split step-$k$ sweep shows max-gradient precision stabilizing from step $\sim 25$ onward, while loss-at-step-$k$ is non-monotonic: loss is stronger at step $11$ for extreme high-loss configurations but trails max-gradient at steps $25$--$100$ (App.~\ref{app:stepk-precision-sweep}). Cross-family thresholds do not transfer; calibration is per family, not a global constant.
\pagebreak[4]

\textbf{DLM-LoRA triage workflow.} The audited workflow is three steps: \textbf{drop} top-$1$ as a PEFT alarm at this horizon, \textbf{log} max-gradient by step $\sim25$, and \textbf{calibrate} thresholds per DLM family before routing high-gradient configurations to inspection or separately validated follow-up sweeps. Three findings follow from existing data: top-$1$ is not a PEFT warning at this horizon, max-gradient is a family-local inspection trigger inside LLaDA-family runs, and mask ratio should be tuned per model rather than exported as a single operating window. Mask ratio is the strongest tested low-cost covariate in the mask-ratio holdout probes; max-gradient is the supported early inspection signal while preserving mask-ratio design as a separate tuning axis.

\paragraph{Contributions.}
\begin{enumerate}
\setlength{\itemsep}{0pt}
\setlength{\parsep}{0pt}
\setlength{\topsep}{2pt}
\item An $816$-configuration refutation: top-$1$ fires in $816/816$ DLM PEFT configurations while $0/816$ actual collapses occur (\S\ref{sec:collapse-metric}).
\item A two-level saturation characterization showing why the warning fails, with token-side pre-equilibrium saturation and parameter-side gradient routing evidence (\S\ref{sec:mechanism}).
\item A family-calibrated max-gradient triage protocol with held-out precision $0.68$ (CI $[0.500,0.947]$) on the pooled LLaDA-family corpus (\S\ref{sec:collapse-metric}).
\item Thirteen falsification probes and matched AR controls that bound the claim to short-horizon DLM-LoRA PEFT (App.~\ref{app:falsification-audit}).
\end{enumerate}
Manuscript values are source-mapped through local run manifests, claim-bearing aggregates, and verification summaries; public paper source, reference scripts, and the sanitized aggregate result artifacts that back the tables and figures are released at \repourl{} (\dataurl{}).

\section{Background}
\label{sec:background}

\subsection{Discrete Diffusion Language Models}
\label{sec:dlm-background}

Discrete diffusion language models (DLMs) train by adding discrete noise to token sequences (masking tokens at rate $\rho$) and learning to reconstruct the original tokens from the noisy input. At inference, DLMs iteratively denoise a fully masked sequence over $T$ steps, using bidirectional attention at each step \citep{llada2025, mdlm2024}. We write the masked-diffusion training objective in the per-masked-token form used by our implementation:
\begin{equation}
    \mathcal{L}(\theta) =
    -\mathbb{E}_{t, \mathbf{x}_t}\left[
    \frac{1}{|\mathcal{M}_t|}
    \sum_{i\in\mathcal{M}_t}
    \log p_\theta(x_0^i \mid \mathbf{x}_t)
    \right],
    \label{eq:mdlm-loss}
\end{equation}
where $\mathcal{M}_t=\{i:x_t^i=\text{\masktoken{}}\}$ and $x_0^i$ is the clean token at position $i$. This differs fundamentally from AR next-token prediction. The density of gradient signal scales with $\rho$ and the prediction entropy grows with the number of tokens jointly predicted, which together drive the rank--mask interaction we characterize in Sec.~\ref{sec:experiments}.\footnote{We encountered five silent-failure modes in the standard HuggingFace + PEFT stack when running LoRA on LLaDA/Dream (loss API returning \texttt{None}, generation API kvcache assertion, target-module auto-detection, Dream model-class loader, Dream attention-mask dtype). Drop-in fixes appear in Appendix~\ref{app:api-code}; public release artifacts are linked in Appendix~\ref{app:reproducibility}.}

\section{Methodology}
\label{sec:methodology}

\subsection{Correct Training Objective}

Standard HuggingFace PEFT training assumes a model-internal supervised loss, but LLaDA-style DLM forward passes return logits only because the caller defines the masking distribution. Following \citet{mdlm2024}, we mask tokens externally and use Eq.~\ref{eq:mdlm-loss} with loss computed only over masked positions. Appendix~\ref{app:api-code} gives the drop-in API fixes needed to reproduce this objective.

\subsection{Experimental Setup}

\paragraph{Models.} We evaluate LoRA fine-tuning in three roles. LLaDA-family DLMs provide the primary top-1 refutation and max-gradient separation; Pythia/Qwen causal models under matched masked-CE serve as diagnostic controls; Dream, MDLM-OWT, and LLaDA-MoE runs act as boundary cohorts. The primary DLM setup is:
\begin{itemize}
    \item \textbf{\llada{}-Instruct} \citep{llada2025}: 8B parameter masked diffusion LM. Mask token ID: 126336. Architecture: LLaDAModel (custom, non-HF-standard).
    \item \textbf{LLaDA2.0-mini} \citep{llada2_2025}: 15.93B MoE masked diffusion LM. Mask token ID: 156895. This model provides the 60-configuration rank$\times$mask surface and the 2$\times$2 task-performance factorial.
    \item \textbf{\dream{}} \citep{dream2025}: 7B parameter masked diffusion LM. Loaded via \texttt{AutoModel}. Requires boolean attention mask.
\end{itemize}

\paragraph{LoRA configuration.} All primary DLM runs adapt attention projections only (\texttt{q\_proj, k\_proj, v\_proj, o\_proj}). The LLaDA2.0-mini surface uses ranks $\{4,8,16,32,64\}$ and 12 mask ratios spanning $\rho\in[0.05,0.95]$; the task-performance factorial uses ranks $\{4,64\}$ and masks $\{0.40,0.90\}$ for 3 seeds per configuration, and the operating-cell method comparison (App.~\ref{app:method-holdout}) uses $n{=}10$ seeds per method at the learning rate selected by the $\alpha$-sweep. The older LLaDA-Instruct pilot uses the coarser $5\times4$ grid $\rho\in\{0.3,0.5,0.7,0.9\}$, and the Dream-7B pilot uses a learning-rate-resolved rank$\times$mask grid described in Appendix~\ref{app:dream}.

\paragraph{Training.} Short pilot runs use 30--40 steps for API validation; the LLaDA2.0-mini surface uses 200 steps at lr$=10^{-4}$ (reported as an observed-prefix diagnostic because the legacy top-1 detector early-stops all 60 traces at step 11). The $2\times2$ factorial uses 1000 steps on a 152-example hand-written arithmetic corpus with 20 held-out prompts and is reported as masked-CE convergence evidence; generation-quality evaluation is separated from this diagnostic claim. Batch size 4, AdamW, gradient norms recorded pre-clipping ($0.5$ threshold). LLaDA2.0-mini runs on \hworkstation; pilots on \hpccluster{} NVIDIA L40S (48GB). Implementation: HuggingFace \texttt{transformers}~\citep{transformers-library} + PEFT~\citep{peft-library}. Gradient norms are global $\ell_2$ norms over trainable LoRA parameters. Family-canonical hyperparameters are used across model families (LLaDA lr$=10^{-4}$ eff.\ batch $64$; Dream-default lr$=2{\times}10^{-6}$; MDLM-OWT lr$=10^{-4}$ batch $1$); the $816/816$ fire-rate identity is an empirical aggregate under these family-specific settings, not a hyperparameter-invariance proof (App.~\ref{app:mdlm-top1}).

\paragraph{Scope note on AR baselines.} Two AR controls play distinct roles. \textbf{Training-stack sanity:} a Mistral-7B LoRA baseline under standard next-token cross-entropy (Appendix~\ref{app:reproducibility}) verifies that the implementation itself is not the instability source. \textbf{Masked-CE control:} Pythia-1B~\citep{biderman2023pythia} on the same $5\times12$ grid (180 runs, $n{=}3$ seeds, \S\ref{sec:ar-control-body}) tests the loss-vs-architecture confound by holding the loss fixed while varying architecture and pretraining; Qwen3.5-9B~\citep{qwen3_2025} adds a larger matched control in \S\ref{sec:ar-control-body}.

\section{Experiments and Results}
\label{sec:experiments}

The experiments answer a diagnostic question, not a method-comparison question: can a low-cost monitor identify DLM-LoRA configurations that should be inspected before the late training loss is known? We first test the transferred top-$1$ warning, then evaluate max gradient norm under the same held-out label, and finally use the LLaDA2.0-mini rank$\times$mask surface, AR controls, token/gradient measurements, and task probe to mark the boundary of the claim. Preliminary \llada{}-Instruct experiments across a $5 \times 4$ rank$\times$mask grid motivated the denser LLaDA2.0-mini study but are omitted from the main body.

\paragraph{Evaluation object and baselines.} The object under evaluation is the \emph{training monitor}. The main baselines are therefore diagnostic: the transferred top-$1$ warning, max-gradient-up-to-step-$k$, loss-at-step-$k$, mask-ratio covariates, and matched AR masked-CE controls. PEFT variants enter as boundary and method-comparison cohorts, but the claim-bearing question stays fixed: whether an early DLM-LoRA monitor can route top-decile final-loss configurations to inspection better than the transferred top-$1$ warning under family-specific thresholds. Table~\ref{tab:monitor-verdict} summarizes the action-facing verdict.

\subsection{Top-1 Has Zero Precision; Max Gradient Norm Provides Calibrated Triage}
\label{sec:collapse-metric}

The audited LLaDA-family runner exposes a top-1-frequency collapse heuristic: it emits \texttt{top1\_warning\_detected} when more than $50\%$ of predicted argmax tokens concentrate on a single token within a short observation window. This makes it a plausible but unvalidated short-run PEFT stability monitor; the Dream and MDLM boundary cohorts use harmonized logging fields for the same test.

\paragraph{Denominator and result.} We aggregate across three DLM model families totalling $\mathbf{816}$ configurations: LLaDA family ($n{=}671$; four cohorts from 2--3 model checkpoints), a Dream-$7$B dense boundary cohort ($n{=}100$; App.~\ref{app:scale-boundary}), and an MDLM-OWT $130$M dense boundary cohort ($n{=}45$; App.~\ref{app:mdlm-top1}). The cohorts use harmonized logging fields for \texttt{top1\_warning\_detected} and post-hoc \texttt{collapsed}; MDLM measures top-$1$ from the training-time masked-input forward pass, while LLaDA-family runs use the corresponding runner proxy (App.~\ref{app:mdlm-top1}). The top-1 collapse warning fires in $\mathbf{816/816}$ ($\mathbf{100\%}$) configurations; actual collapse occurs in $\mathbf{0/816}$ ($\mathbf{0\%}$). The diagnostic has zero precision at this horizon. In PEFT fine-tuning at $\leq$200 steps, the warning fires on a pre-equilibrium artifact of LoRA updates rather than on divergence dynamics. AR controls ($0/360$ collapses across the audited main and extended-mask grids; \S\ref{sec:ar-control-body}) show that the tested masked-CE controls do not produce an analogous collapse pattern. Source-level provenance is recorded in Appendix~\ref{app:reproducibility}.

\paragraph{Max gradient norm separates stable from unstable configurations.} We report the maximum LoRA gradient $\ell_2$ norm over the training trajectory (pre-clipping at the standard $0.5$ threshold) as the triage signal. Within DLM family, the median max-gradient ratio between unstable (top-decile final-loss) and stable (sub-median final-loss) configurations is $\mathbf{3.23\times}$ on LLaDA2.0-mini full surface ($n{=}144$, Mann--Whitney $U$, $p_{\text{Bonf}}{=}2.7\times10^{-7}$, $m{=}6$, bootstrap $95\%$ CI $[2.76, 3.97]$); $\mathbf{362\times}$ on the LLaDA method-comparison set ($n{=}395$; stable median $99.3$, unstable median $35{,}960.4$ in the source scale; $p_{\text{Bonf}}{=}5\times10^{-21}$, CI $[202, 779]$); and $1.48\times$ ($p_{\text{Bonf}}{=}0.036$) on the 10-seed critical expansion ($n{=}120$, compressed dynamic range). AR controls show smaller, inconsistent separation (Table~\ref{tab:hero-top1-maxgrad-summary}; verification summaries in App.~\ref{app:reproducibility}), supporting family calibration rather than a global threshold.

\paragraph{Held-out precision check.} A fixed $80/20$ split over the full $671$-configuration LLaDA-family corpus ($n_{\text{train}}{=}536$, $n_{\text{test}}{=}135$), including the method-comparison cohort ($n{=}395$), with the max-gradient threshold $F_1$-optimized on train predicts top-decile final-loss on test with precision $\mathbf{0.68}$, recall $\mathbf{0.94}$, and $F_1{=}\mathbf{0.79}$.\footnote{The top-decile final-loss label is defined on the full corpus; by chance the test-split unstable fraction is $13.3\%$ ($18/135$) vs.\ $9.3\%$ in train.} A separate $B{=}1000$ split-bootstrap, which resamples configurations, redraws the train/test split, and reselects the threshold on train in each replicate, gives $95\%$ precision CI $[\mathbf{0.500}, \mathbf{0.947}]$, disjoint from the always-positive baseline ceiling $0.148$. This is a roughly $5\times$ precision lift over the fixed-split baseline; even the lower confidence bound is more than $3\times$ the baseline ceiling, but the absolute precision remains moderate. The supported use is therefore inspection and routing rather than an automatic decision rule; the pooled evaluation is appropriate because the primary grid alone is underpowered for threshold calibration (Limitations). A late-vs-early gradient-ratio rule and its conjunction with the threshold are less precise (App.~\ref{app:reproducibility}); the next paragraph reports a separate $B{=}200$ random-split timing sweep.

\begin{figure}[t]
\centering
\includegraphics[width=\columnwidth]{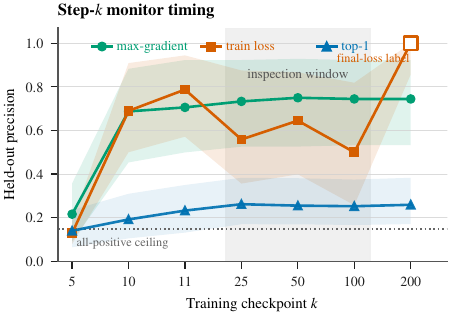}
\caption{\textbf{Step-$k$ precision separates stable inspection from final-loss hindsight.}
On the $671$-configuration LLaDA-family corpus, max-gradient precision stabilizes through the step-$25$--$100$ window where loss-at-step-$k$ is least reliable; the step-$200$ loss point is the final-loss label. Colored ribbons are bootstrap $95\%$ CIs; the gray band marks the inspection window; the dotted line is the all-positive precision ceiling. The top-$1$ line (blue) exceeds this ceiling at $k{\geq}10$ because random splits can correlate top-$1$ with loss by chance; the zero-precision result is on the fixed full-corpus split ($0/816$ collapses).}
\label{fig:stepk-precision}
\end{figure}

\paragraph{Early-warning timing: stable inspection before late loss settles.} The practical case for max gradient norm is timing, not absolute precision or compute saving. We sweep three predictors --- max-gradient-up-to-step-$k$, loss-at-step-$k$, and max top-$1$ token-frequency-up-to-step-$k$ --- across $k\in\{5,10,11,25,50,100,200\}$ on the same $671$-configuration corpus with $B{=}200$ random $80/20$ splits and $F_1$-optimized thresholds on train (Fig.~\ref{fig:stepk-precision}; Tab.~\ref{tab:stepk-precision-sweep}). Max-gradient precision stabilizes at $\mathbf{0.73}$--$\mathbf{0.75}$ from step $\mathbf{25}$ onward. Loss-at-step-$k$ is non-monotonic: it spikes to $0.79$ at step $11$, dips to $0.50$--$0.65$ at steps $50$--$100$, and becomes tautological at step $200$ because loss is then the label. A practitioner reading only loss at step $11$ would observe higher single-point precision ($0.79$) but loses reliable signal for any inspection triggered between steps $25$ and $100$; max gradient norm accumulates in the same training logs without requiring an additional forward pass and remains predictive across the full step-$25$--$100$ window. Max top-$1$-up-to-step-$k$ never exceeds $0.27$, consistent with the pre-equilibrium-artifact framing.

\paragraph{DLM-LoRA triage workflow.} The operational recipe is deliberately narrow. \textbf{Drop} the top-$1$ collapse warning as a PEFT early-warning signal at $\leq 1$K step horizons. \textbf{Log} max LoRA gradient norm at every training step and inspect it by step $\sim25$. \textbf{Calibrate} high-gradient thresholds per family, e.g., absolute max-grad values above $50$--$100$ in the LLaDA2.0-mini logging scale or above a locally calibrated high quantile. These thresholds are inspection triggers, not prospectively validated cross-family cutoffs or compute-saving policies. Appendix~\ref{app:reproducibility} gives the logging fields needed to implement this protocol.

\begin{table}[t]
\centering
\caption{\textbf{Actionable monitor verdict.} The deliverable is not a new PEFT method; it is a claim-matched triage protocol for DLM-LoRA training monitors at the tested horizons.}
\label{tab:monitor-verdict}
\footnotesize
\setlength{\tabcolsep}{3pt}
\begin{tabular}{@{}p{0.28\columnwidth}p{0.64\columnwidth}@{}}
\toprule
\textbf{Decision} & \textbf{Evidence-backed use} \\
\midrule
\textbf{Drop warning} & Top-$1$ fires in $816/816$ DLM configurations with $0/816$ actual collapse; not a PEFT collapse detector at this horizon. \\
\textbf{Use triage} & Fixed-split max-gradient precision $0.68$, recall $0.94$; $B{=}200$ random-split sweep stable by step $25$; inspection trigger only. \\
\textbf{Keep baseline} & Loss-at-$k$ precision reaches $0.79$ at step $11$ but falls to $0.50$--$0.65$ at steps $50$--$100$; step $200$ is the label by construction. \\
\textbf{DLM-LoRA only} & AR masked-CE controls have $0/360$ collapse and smaller or inconsistent separation, so the warning failure is scoped to the tested DLM-LoRA monitor transfer. \\
\textbf{No global cutoff} & Cross-family thresholds do not transfer; high-gradient values are inspection triggers only after per-family calibration. \\
\bottomrule
\end{tabular}
\end{table}

\subsection{\texorpdfstring{A U-Shaped Gradient Instability Profile Across Mask Ratio}{A U-Shaped Gradient Instability Profile Across Mask Ratio}}
\label{sec:llada2-surface}

We extend the analysis to \textbf{LLaDA2.0-mini}~\citep{llada2_2025} (inclusionAI/LLaDA2.0-mini, 15.93B MoE, mask token ID 156895), a more recent and larger masked diffusion model. We configure 60 unique rank$\times$mask combinations ($n{=}144$ total runs including multi-seed replications): ranks $\{4,\,8,\,16,\,32,\,64\}$ $\times$ 12 mask ratios spanning $\rho \in [0.05, 0.95]$ (raw grid in Appendix Table~\ref{tab:llada2-surface}), with a 200-step budget at lr $=10^{-4}$. The legacy top-1 collapse detector early-stops all 60 raw traces at step 11, so this surface should be read as an observed-prefix short-run diagnostic rather than a completed 200-step trajectory.

\paragraph{U-shaped instability profile.} Unlike the non-monotone rank-optimum flip observed in LLaDA-Instruct (\S\ref{sec:experiments}), LLaDA2.0-mini reveals a \textbf{U-shaped gradient instability profile} across mask ratio (companion to Figure~\ref{fig:hero-top1-maxgrad} panel B). Because top-$1$ fires uniformly, this surface explains where max-gradient triage from \S\ref{sec:collapse-metric} becomes useful. The high-mask arm corresponds to \textbf{high-mask gradient amplification}: in this observed-prefix grid, fine-tuning LoRA on a DLM at $\rho > 0.70$ produces gradient magnitudes up to $6.0{\times}$ larger than the operating-window maximum (34.8 vs.\ 5.8), in proportion to LoRA rank. The left arm is sparse-signal variance. We use these two names throughout:

\textbf{Two instability mechanisms:}
\begin{itemize}
    \item \textbf{Left arm} (mask $< 0.15$): Sparse supervision, only 5--15\% of tokens are masked per sequence. The per-batch gradient estimate has high variance (few prediction targets, noisy signal). Gradient norm spikes reach 7.7--23.6 across ranks.
    \item \textbf{Right arm} (mask $> 0.70$): \textbf{High-mask gradient amplification}, predicting 70--95\% of tokens simultaneously produces a high-entropy prediction task with large loss and gradient magnitudes. Rank amplifies this arm directionally: in the 1-seed surface (Table~\ref{tab:llada2-surface}) $r{=}4$ at $\rho{=}0.95$ reaches 2.7 and $r{=}64$ reaches 34.8 ($12.9\times$); in the 3-seed replication (Table~\ref{tab:llada2-3seed-surface}) the same configurations give $34.5{\pm}9.7$ and $41.2{\pm}17.5$ respectively ($1.19\times$, $n{=}3$ seeds), so we keep the high-mask asymmetry as a directional finding and the 3-seed values as the canonical magnitude.
    \item \textbf{Low-mid operating region} (mask $\in [0.30, 0.40]$): in the direct one-seed observed-prefix grid, these configurations have low gradient norms across all five ranks, with $\rho{=}0.45$ supported only by a narrower $r{=}64$ boundary run. In the 3-seed completed grid, the lowest mean gradient norms shift toward low-mid masks and the $r{=}64$ values at $\rho\in\{0.30,0.40\}$ are noisy; we therefore base the practical recommendation on convergence and held-out CE evidence rather than on a replicated global gradient minimum. \textbf{Practical recommendation: avoid $\rho>0.70$ for LLaDA2.0-mini LoRA at lr=$10^{-4}$ in the tested setup; treat $\rho=0.30$--$0.40$ as a conservative low-mid default, not a global optimum.}
\end{itemize}

LLaDA2~\citep{llada2_2025} independently reports high gradient variance at extreme masking during pre-training and clips their noise-schedule coefficient within $[\alpha_{\min}, \alpha_{\max}]$, a bandwidth that maps to our operating window. Our LoRA characterization adds the rank dimension and shows amplification concentrates on the high-mask arm.

\paragraph{Replication correction.} The 1-seed observed-prefix $12.9\times$ high-mask amplification contracts to $1.19\times$ under the 3-seed full-grid replication (Table~\ref{tab:llada2-surface}), with high-mask std as large as $60$. We therefore use the replicated surface as the canonical magnitude estimate and keep symbolic-regression descriptors as appendix-only exploratory summaries, not decision rules.

\paragraph{Standard AR PEFT does not traverse this surface.} Standard AR LoRA fine-tuning uses dense next-token supervision over all non-first positions, has no mask-ratio dimension, and therefore cannot exhibit a rank--mask interaction in its standard recipe. The natural follow-up question is whether the U-shape we observe on LLaDA2.0-mini is a property of DLM bidirectional architecture or of the masked-CE objective itself. We answer that with a paired AR control in the next subsection.

\subsection{AR Baseline Control (summary; full grids in App.~\ref{app:ar-control})}
\label{sec:ar-control-body}

To isolate the masked-CE objective from DLM bidirectional attention, we ran matched random-mask cross-entropy controls on Pythia-1B, Pythia $\{410\text{M},2.8\text{B},6.9\text{B}\}$, and Qwen3.5-9B, with $360$ audited configurations across main grids and extended-mask supplements. At matched configuration $(r{=}64, \rho{=}0.40)$, the max-grad-norm magnitude is $\mathbf{2.0\times}$--$\mathbf{2.5\times}$ smaller on AR than on LLaDA2.0-mini ($16.61$ on Pythia-1B vs $33.4$ on LLaDA2.0-mini); the high-vs-mid ratio at $r{=}64$ is $1.20\times$ on Pythia-1B AR control vs $2.54\times$ on LLaDA2.0-mini, and Qwen3.5-9B shows a cross-family mid-mask peak rather than the U-shape. AR controls report $0/360$ actual collapses, supporting a DLM-family-scoped interpretation rather than a masked-CE-generic one. The grid, denominator, and cross-architecture detail live in App.~\ref{app:ar-control}; we keep here only the body-essential conclusion: masked-CE alone is not sufficient to reproduce the DLM-family rank-amp magnitude, so the max-gradient triage signal in \S\ref{sec:collapse-metric} is calibrated against the DLM family it serves.

\subsection{DLM Scale-Architecture Boundary (summary; full grids in App.~\ref{app:scale-boundary})}
\label{sec:cross-arch-summary}
\label{sec:cross-arch}

The LLaDA2.0-mini operating window survives unevenly across DLM scales and architectures. The loss-side high-mask disadvantage replicates on Dream-7B (7B dense, lr-calibrated; App.~\ref{app:dream}) and LLaDA2.1-mini (4-configuration transfer; App.~\ref{app:llada21-transfer}), but the rank-amplification direction is mixed on MDLM-OWT-130M ($n{=}3$ replication; the earlier single-seed lr-modulation pattern does not survive replication) and softens on LLaDA-MoE-A1B (1.4B small-MoE: gradient-side amplification $1.89$--$2.47\times$, loss-side flat; App.~\ref{app:moe-scale-probe}). We claim DLM-family scope rather than an architecture-general window, with per-model lr calibration required; the full scale-boundary table and per-model lr discussion are reported in App.~\ref{app:scale-boundary}.

\subsection{Why Top-1 Fires in Every DLM Configuration: A Two-Level Characterization}
\label{sec:mechanism}
\label{sec:theory}

The $816/816$ fire vs $0/816$ collapse asymmetry reflects a structural mismatch between what the metric measures and what training stability requires. We characterize it with two corpus-wide measurements that decouple token-space concentration from parameter-space gradient routing.

\paragraph{Level 1 (token-side): top-1 is saturated before training.}
Across all $671$ LLaDA-family configurations, the top-$1$ token frequency at training step $0$ has mean $0.83$ and standard deviation $0.13$; $100\%$ of configurations are already above $0.5$ at step $0$, and $65\%$ are already above $0.8$. The median configuration crosses $0.95$ within $\mathbf{4}$ optimizer steps. The legacy detector samples at step $11$ and fires in every configuration because the threshold ($0.5$) is below the corpus-wide initialization distribution. Stable configurations (sub-median final loss) and unstable configurations (top-decile final loss) have indistinguishable median fire-step ($11$ vs $11$; Mann--Whitney $U$ two-sided $p{=}0.20$, $n_{\text{stable}}{=}336$, $n_{\text{unstable}}{=}68$; remaining $267$ mid-band configurations are excluded from this stability contrast). The complementary saturation-step diagnostic (first step where top-$1$ crosses $0.95$) is significant in the \emph{opposite} direction: unstable configurations saturate \emph{faster} (median $1.0$) than stable configurations (median $4.0$), $p{=}4.7\times10^{-5}$ ($n_{\text{stable}}{=}178$, $n_{\text{unstable}}{=}68$; conditioned on configurations that crossed $0.95$ by step $200$). A signal that saturates before training, fires faster on unstable runs, and is uniform across the corpus cannot discriminate stability; it measures a pre-equilibrium argmax-concentration artifact of LoRA's small-magnitude initialization plus a few masked-CE updates against an already-confident pre-trained DLM. Figure~\ref{fig:preequilibrium-top1} reports the aggregate timing evidence.
\paragraph{Level 2 (parameter-side): rank-amp is optimization routing, not token routing.}
The token-side concentration above coexists with a near-uniform per-position information density, so the warning signal must measure something other than token-distribution concentration. At the worst rank-amplification corner ($r{=}64$, $\rho{=}0.95$, LLaDA2.0-mini, $n{=}3$ seeds, last-$10$ steps; App.~\ref{app:parameter-routing}), the per-token cross-entropy-gradient distribution has Gini $0.287\pm0.056$ and the largest evaluated token position contributes only $1.54\%\pm0.17\%$ of total CE-gradient mass (uniform baseline $0.8\%$). In the same runs, the LoRA-parameter gradient distribution has Gini $0.463\pm0.031$ and a \emph{single} LoRA matrix carries $63.0\%\pm3.6\%$ of total parameter-side gradient mass. At this high-mask corner, rank-amplification is therefore an \emph{optimization-routing} phenomenon: the masked-CE signal arrives spread across token positions but is funnelled through a small subset of high-rank LoRA adapters in the late trajectory. Max gradient norm samples that late-trajectory routing, supporting why it carries discriminative information that top-$1$ does not.

\paragraph{What this characterization predicts.}
The useful monitor should depend on late-trajectory parameter dynamics, not early token confidence: max-gradient fits this pattern inside the calibrated LLaDA-family split, while the always-positive top-$1$ warning does not. The pathology is scoped to LoRA-on-pretrained-DLM regimes; AR controls (App.~\ref{app:ar-control}) and the DLM scale-boundary check (App.~\ref{app:scale-boundary}) support this boundary. \citet{mdlm-implicit-reg2026} studies masked-diffusion signal/noise decomposition in a different generalization regime.

The full pre-equilibrium trajectory and timing breakdown are shown in Fig.~\ref{fig:preequilibrium-top1}.

\paragraph{Why no single-axis intervention prevents saturation.}
\label{sec:saturation-theorem-body}
The empirical $816/816$ identity is consistent with a masked-CE convergence argument: if fitting increases expected top-$1$ mass before optimization settles, then convergence-preserving single-axis interventions should preserve the legacy fire event. Two probes show the boundary. A loss-level entropy bonus on MDLM-OWT with $\lambda\in\{0.5,1.0,2.0,5.0,10.0\}$ does not reduce top-$1$ mass at this horizon ($-0.008$ at $\lambda{=}0.5$, $+0.051$ at $\lambda{=}10$; App.~\ref{app:entropy-bonus-probe}); canonical PiSSA improves MDLM-OWT final loss ($-0.43$ at $200$ steps; $1.82\to1.24$, paired delta $-0.57$ at step $1000$) without changing the fire identity (App.~\ref{app:spectral-init-canonical}). App.~\ref{app:falsification-audit} reports all thirteen probes; the bound is explanatory scaffolding, not a load-bearing theorem.

\paragraph{Scope refinements.}
\label{sec:scope-refinements}
The low-mid operating region does not define an architecture-general optimum: some final-loss probes prefer lower masks, while the convergence and held-out CE probes mainly support avoiding high-mask regimes in LLaDA-family settings (App.~\ref{app:reproducibility}). The worst rank-amplification corner shifts from $\rho{=}0.95$ ($12.9\times$, one seed) to $\rho{=}0.90$ ($84.7{\pm}60.4$, three seeds); high-mask capacity effects and LLaDA2.1-mini transfer remain underpowered. We therefore state a scoped diagnostic, not an architecture-general recipe.

\paragraph{Task-performance sanity check.}
\label{sec:task_perf}
A small in-domain masked-CE convergence probe checks whether the gradient surface predicts downstream loss reduction, not generation accuracy. LLaDA2.0-mini is trained for $1000$ steps on $152$ hand-written arithmetic examples, crossing rank $\{4,64\}$ with mask ratio $\{0.40,0.90\}$ and evaluating masked-CE on $20$ disjoint prompts (App.~\ref{app:holdout-prompts}).

\noindent\textbf{Finding.} Table~\ref{tab:task_perf} matches the surface ordering: operating-window configurations ($\rho{=}0.40$) reach lower final and holdout losses than high-mask configurations ($\rho{=}0.90$). The within-window rank gap is not significant (paired $t$ $p{=}0.40$), and Table~\ref{tab:multibench-masked-ce} shows no Bonferroni-corrected rank-64 advantage.

\FloatBarrier
\section{Related Work}
\label{sec:related}

LoRA/PEFT work introduces low-rank and quantized adapters \citep{lora2021,qlora2023,dora2024} plus rank-allocation and optimizer-side variants \citep{galore2024,sensitivity-lora2025,lora-mgpo2025,riemannian-lora2025}, but these works study generic or AR adaptation regimes rather than DLM mask-ratio monitor transfer. DLM work studies objectives and decoding \citep{mdlm2024,llada2025,dream2025}, scaling and surveys \citep{llada2_2025,dlm-survey2025}, mask-agnostic fine-tuning \citep{piskorz2025masks}, and recent systems or adapters including noise-aware LoRA \citep{lad2025,dare2026,gift2025,wang2026nara}; these improve dLLM adaptation or inference but do not test whether top-$1$ collapse warnings transfer into supervised LoRA fine-tuning. We use the term collapse for training-time top-$1$ argmax saturation, distinct from the representational layer collapse reported in fully-trained DLMs \citep{layercollapse2026}; auditing warning-signal precision under matched false-positive control has precedent outside language modeling \citep{recursivecollapse2026}. Like \citet{schaeffer2023emergent}, we show that a familiar metric changes meaning outside its calibration regime; App.~\ref{app:extended-related} gives the fuller taxonomy.

\FloatBarrier
\section{Conclusion}
\label{sec:conclusion}

Top-$1$ fires in $816/816$ configurations while observed collapse is $0/816$ across three DLM families because the token-side signal saturates before training stability is observable. Max gradient norm instead gives a family-local inspection signal: precision $0.68$ on the pooled LLaDA-family split and stable step-$25$--$100$ behavior. The scoped recommendation is to drop top-$1$ as a PEFT collapse warning, log max-gradient for inspection, and recalibrate mask ratio per model before reusing inference-time confidence monitors as training alarms.

\section*{Limitations}
\label{sec:limitations}

\textbf{Budget and seeds.} The primary $60$-configuration rank$\times$mask grid uses $n{=}3$ seeds at $200$ training steps, expanded to $n{=}10$ at twelve critical configurations. Power analysis (App.~\ref{app:bootstrap-rank-amp}) places adequate detection of $2\times$ ratios at $n\geq 30$ for high-mask configurations, so rank-amplification magnitudes are directional estimates; the top-$1$ refutation and max-gradient triage claims rest on the larger audited denominators.

\textbf{Architecture and adapter scope.} The max-gradient precision claim is calibrated on the pooled LLaDA-family corpus ($n{=}671$); the primary rank$\times$mask grid alone ($n{=}264$) contains too few unstable configurations to calibrate a held-out threshold reliably, so the pooled evaluation is the appropriate unit. The zero-precision top-$1$ denominator additionally includes Dream-$7$B and MDLM-OWT-$130$M boundary cohorts across four LLaDA-family cohorts (2--3 model checkpoints). Adapters are placed on attention projections (\texttt{q,k,v,o}); MLP, embedding, and LM-head LoRA placement, quantization-mask interaction~\citep{quant_dllm_2026, fast_dllm_2026}, and fully matched dense LLaDA-$8$B replication are follow-up axes.

\textbf{Task and use scope.} The task probe is an in-domain masked-CE convergence check rather than an accuracy-grade generation benchmark. Max-gradient is therefore presented as the tested LLaDA-family alternative to top-$1$ for early inspection, while coupled $(\rho,r,\text{family})$ intervention design and generation-quality gains remain separate claims for future work. Low-mid mask ratios are a conservative LLaDA-family default in the tested setup, not a global optimum; per-architecture validation is required.

\textbf{Diagnostic horizon.} The $816$/$816$ zero-precision result is bounded to short-run PEFT diagnostics at the tested horizon. We test the inherited legacy warning threshold ($>50\%$ argmax concentration at step $11$); recalibrated thresholds or alternative top-$1$-derived statistics could behave differently and remain unvalidated. Separate $2000$-step sidecars on Dream-$7$B ($27/27$ fire, $0/27$ collapse) and LLaDA2.0-mini MoE ($9/9$ fire, $0/9$ collapse) are consistent with this warning-failure pattern, but remain outside the $816$-configuration headline denominator. The result should not be read as a claim about full fine-tuning, DLM pretraining from scratch, or budgets beyond these bounded sidecars.

\section*{Ethical considerations}
\label{sec:ethics}

All training and evaluation data are publicly released English-language benchmarks under permissive licenses (GSM$8$K~\citep{gsm8k2021}, HumanEval~\citep{humaneval2021}, MMLU~\citep{mmlu2021}: MIT; MetaMathQA-$5$K: CC-BY-NC-SA-$4.0$); the $152$-example instruction corpus is hand-written, no PII, no scraped third-party content. Backbone weights are publicly released (LLaDA-family per model cards; LLaDA-MoE-$7$B-A$1$B per \citealp{llada_moe_a1b_2025}; Pythia~\citealp{biderman2023pythia} + Qwen$3.5$-$9$B~\citealp{qwen3_2025} under Apache $2.0$). Aggregate compute is $\sim119$\,kg\,CO$_2$eq total, estimated from reported GPU-hours and US grid-intensity context~\citep{electricitymap2024}. The max-gradient triage protocol operates only on training diagnostics and produces no model outputs; we do not anticipate disproportionate or novel harms beyond those already present in supervised LoRA fine-tuning. AI assistants were used for coding support, layout repair, audit checklists, and prose editing; all claims, numbers, and experimental results were author-verified against local run artifacts. Public artifacts include the arXiv source, reference logging scripts, and sanitized aggregate result JSON/CSV files backing the tables and figures (\repourl{}; \dataurl{}). The public artifacts intentionally exclude raw per-run prompts/completions, W\&B metadata, local paths, checkpoints, and adapter weights.

\bibliography{references}

\clearpage
\appendix
\renewcommand{\thesection}{\Alph{section}}
\makeatletter
\renewcommand\Alph[1]{%
  \ifnum\value{#1}>26 A\@arabic{\numexpr\value{#1}-26\relax}\else\@Alph{\value{#1}}\fi%
}
\makeatother

\input{appendix_focused}
\end{document}

%% file: appendix_focused.tex
\section{Reproducibility and Source Trace}
\label{app:reproducibility}
\label{app:api-code}

\paragraph{Reproducibility scope.}
The arXiv source package contains the manuscript source, bibliography, and rendered figures. The public artifact release contains paper source, reference scripts, and the sanitized aggregate result JSON/CSV files that back the tables and figures at \repourl{} (\dataurl{}). The manuscript values are source-mapped through local run manifests and claim-bearing aggregates rather than copied from tracker prose. The release excludes raw prompts/completions, W\&B metadata, local paths, checkpoints, and adapter weights.

\paragraph{Compute and setup.}
The primary LLaDA-family experiments use H100 NVL-class GPUs; Dream, preliminary LLaDA, and some AR controls use L40S-class GPUs. The paper accounts for approximately $396$ GPU-hours across reported experiment groups and estimates $\approx119$ kg CO$_2$eq. Models are used through HuggingFace \texttt{transformers} and PEFT with explicit masked cross-entropy on masked positions. API pitfalls needed for reproduction are: DLM forward passes may not return a supervised loss, \texttt{generate()} is not the training-time denoising loop, target modules must be explicit, Dream-$7$B loads through \texttt{AutoModel}, and Dream attention masks must be boolean.

\begin{table}[!ht]
\centering
\caption{\textbf{Top-$1$ warning and max-gradient summary.} The top-$1$ warning fires in every audited DLM-family configuration while actual collapse is zero at the tested horizon. Max-gradient separation is family-local, not a global threshold.}
\label{tab:hero-top1-maxgrad-summary}
\scriptsize
\setlength{\tabcolsep}{2pt}
\renewcommand{\arraystretch}{0.94}
\begin{tabular}{@{}L{0.43\linewidth}L{0.21\linewidth}L{0.30\linewidth}@{}}
\toprule
\textbf{Cohort} & \textbf{Top-$1$ / coll.} & \textbf{Max-grad evidence} \\
\midrule
DLM LLaDA2.0-mini-full ($n{=}144$) & 144/144; 0/144 & $3.23\times$ $[2.76,3.97]$; $2.7{\times}10^{-7}$ \\
DLM LLaDA2.0-mini-crit12 ($n{=}120$) & 120/120; 0/120 & $1.48\times$ $[1.12, 1.76]$; $0.036$ \\
DLM LLaDA-method-comp ($n{=}395$) & 395/395; 0/395 & $362\times$ $[202, 779]$; $5{\times}10^{-21}$ (source scale) \\
DLM LLaDA2.1-mini ($n{=}12$) & 12/12; 0/12 & small $n$ \\
DLM Dream-7B boundary ($n{=}100$) & 100/100; 0/100 & boundary cohort \\
DLM MDLM-OWT-130M boundary ($n{=}45$) & 45/45; 0/45 & boundary cohort \\
\midrule
AR Pythia/Qwen masked-CE controls ($n{=}360$) & --; 0/360 & smaller or inconsistent \\
\bottomrule
\end{tabular}
\end{table}

\section{Top-1 Saturation and Step-$k$ Precision Sweep}
\label{app:preequilibrium-trajectories}

\begin{figure}[!ht]
\centering
\includegraphics[width=\columnwidth]{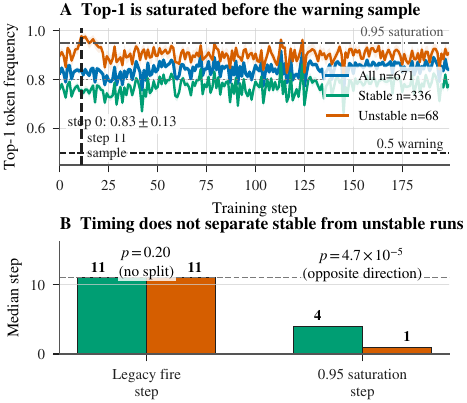}
\caption{\textbf{Top-$1$ collapse is a pre-equilibrium artifact.} (A) Across $671$ LLaDA-family configurations, top-$1$ mass starts high and crosses the legacy threshold before the detector samples. (B) The legacy fire-step has no stable/unstable split, while the stricter $0.95$ saturation step points in the opposite direction: unstable configurations saturate faster.}
\label{fig:preequilibrium-top1}
\end{figure}

\paragraph{Timing.}
Top-$1$ token frequency is $0.83\pm0.13$ at step $0$ on the LLaDA-family corpus. All configurations are already above $0.5$ at step $0$, and the median crosses $0.95$ within four optimizer steps. Stable and unstable configurations have the same median legacy fire-step ($11$ vs. $11$, Mann--Whitney $p{=}0.20$). The stricter $0.95$ crossing points in the wrong direction for a collapse detector: unstable configurations saturate faster.

\subsection{Step-k Precision Sweep}
\label{app:stepk-precision-sweep}

\begin{table}[!ht]
\centering
\caption{\textbf{Step-$k$ held-out precision sweep.} Median precision over $B{=}200$ random $80/20$ splits on $671$ LLaDA-family configurations. Max-gradient is stable from step $25$ onward; step-$200$ loss is the label by construction.}
\label{tab:stepk-precision-sweep}
\footnotesize
\setlength{\tabcolsep}{4pt}
\resizebox{\columnwidth}{!}{%
\begin{tabular}{@{}rccc@{}}
\toprule
$k$ & \textbf{max-grad} & \textbf{loss-at-$k$} & \textbf{max top-$1$} \\
\midrule
$5$   & $0.22$ $[0.09,0.36]$ & $0.13$ $[0.08,0.18]$ & $0.14$ $[0.06,0.24]$ \\
$10$  & $0.69$ $[0.45,0.88]$ & $0.69$ $[0.50,0.91]$ & $0.19$ $[0.11,0.31]$ \\
$11$  & $0.71$ $[0.50,0.92]$ & $0.79$ $[0.57,0.94]$ & $0.23$ $[0.13,0.35]$ \\
$25$  & $0.73$ $[0.53,0.92]$ & $0.56$ $[0.36,0.88]$ & $0.26$ $[0.17,0.38]$ \\
$50$  & $0.75$ $[0.53,0.93]$ & $0.64$ $[0.40,0.87]$ & $0.26$ $[0.17,0.38]$ \\
$100$ & $0.74$ $[0.53,0.92]$ & $0.50$ $[0.26,0.82]$ & $0.25$ $[0.17,0.38]$ \\
$200$ & $0.74$ $[0.53,0.92]$ & $1.00$ $[0.86,1.00]$ & $0.26$ $[0.17,0.38]$ \\
\bottomrule
\end{tabular}}
\end{table}

\section{Surfaces, Controls, and Task Probe}
\label{app:llada2-surface-values}
\label{app:llada21-transfer}
\label{app:scale-boundary}
\label{app:dream}
\label{app:mdlm-top1}
\label{app:moe-scale-probe}
\label{app:ar-control}
\label{app:ci}
\label{sec:ci}
\label{app:bootstrap-rank-amp}
\label{app:holdout-prompts}
\label{app:masking-ablation}
\label{app:snr-eff}

\begin{table}[!ht]
\centering
\caption{\textbf{LLaDA-family surface summaries.} The table preserves the source-traced values used in the body; sanitized per-configuration grids are released in the public artifact.}
\label{tab:llada2-surface}
\label{tab:llada2-3seed-surface}
\label{tab:llada2-10seed-critical}
\label{tab:llada21-transfer-summary}
\scriptsize
\setlength{\tabcolsep}{2pt}
\renewcommand{\arraystretch}{0.94}
\begin{tabular}{@{}L{0.29\linewidth}L{0.54\linewidth}L{0.13\linewidth}@{}}
\toprule
\textbf{Evidence slice} & \textbf{Source-traced contrast} & \textbf{Reading} \\
\midrule
1-seed $r{=}64$ surface & max-grad $\rho{=}0.30/0.40$: $4.6/5.8$; $\rho{=}0.90/0.95$: $15.0/34.8$ & tail \\
r4 replicated rows & mean max-grad (n10/n3): $\rho{=}0.40$: $16.4{\pm}1.3$; $\rho{=}0.90/0.95$: $63.3{\pm}60.0/34.5{\pm}9.7$ & noisy \\
r64 replicated rows & mean max-grad (n10/n3): $\rho{=}0.40$: $33.4{\pm}11.0$; $\rho{=}0.90/0.95$: $84.7{\pm}60.4/41.2{\pm}17.5$ & elevated \\
3-seed rank ratio & $r4{\to}r64$ ratio $\rho{=}0.40$: $2.04\times$; $\rho{=}0.90/0.95$: $1.34/1.19\times$ & corrected \\
10-seed critical configs & $\rho{=}0.90$ vs $0.40$: max-grad ratios $3.87/1.69/2.53\times$ and final-loss deltas $+0.17/+0.17/+0.18$ for ranks $4/16/64$ & desc. \\
LLaDA2.1 transfer & $\rho{=}0.90$ vs $0.40$: max-grad ratios $1.76/3.52\times$ and final-loss deltas $+1.11/+0.99$ for ranks $4/64$ & scoped \\
\bottomrule
\end{tabular}
\end{table}

\phantomsection\label{fig:llada2-surface-heatmap}
\paragraph{Scale boundary.}
The LLaDA2.0-mini low-mid mask recommendation is not architecture-general. Loss-side high-mask disadvantage replicates on Dream-7B after learning-rate calibration and in a small LLaDA2.1-mini transfer check, but rank-amplification direction is mixed on MDLM-OWT-130M and softens on LLaDA-MoE-A1B. This is why the body states a DLM-family diagnostic and requires per-model calibration.
\phantomsection\label{tab:scale-boundary}
\phantomsection\label{tab:dream_surface}

\begin{table}[!ht]
\centering
\caption{\textbf{In-domain convergence probe.} Values are mean $\pm$ std over $3$ seeds; lower held-out CE is better.}
\label{tab:task_perf}
\footnotesize
\setlength{\tabcolsep}{4pt}
\begin{tabular}{@{}llrrr@{}}
\toprule
\textbf{Regime} & \textbf{Rank} & \boldmath$\rho$ & \textbf{Max$\|\nabla\|$} & \textbf{Holdout CE} \\
\midrule
stable & 4  & 0.40 & 18.0 & $0.43\pm0.08$ \\
stable & 64 & 0.40 & 21.0 & $0.42\pm0.10$ \\
high mask & 4  & 0.90 & 73.9 & $0.57\pm0.06$ \\
high mask & 64 & 0.90 & 96.4 & $0.48\pm0.04$ \\
\bottomrule
\end{tabular}
\end{table}

\begin{table}[!ht]
\centering
\caption{\textbf{Operating-window multi-benchmark masked-CE check.} The $\rho{=}0.40$ rank contrast is not significant after correction. These null results bound the low-mid-mask recommendation to the DLM-LoRA training diagnostic and do not support a downstream generation-quality claim.}
\label{tab:multibench-masked-ce}
\scriptsize
\setlength{\tabcolsep}{3pt}
\resizebox{\columnwidth}{!}{%
\begin{tabular}{lrrrrr}
\toprule
\textbf{Benchmark} & \textbf{n} & \textbf{r4 CE} & \textbf{r64 CE} & \boldmath$\Delta$ & \boldmath$p_{\mathrm{Bonf}}$ \\
\midrule
GSM8K-test & 1319 & $2.66{\pm}2.30$ & $1.92{\pm}0.32$ & $+0.74$ & 1.00 \\
HumanEval & 164 & $1.10{\pm}0.03$ & $1.64{\pm}0.31$ & $-0.54$ & 1.00 \\
MMLU-subset & 250 & $1.10{\pm}0.04$ & $1.27{\pm}0.11$ & $-0.17$ & 1.00 \\
\bottomrule
\end{tabular}}
\end{table}

\phantomsection\label{tab:ci}
\paragraph{AR control.}
Pythia and Qwen masked-CE controls show $0/360$ actual collapses and smaller or inconsistent max-gradient separation. The denominator is Pythia-1B main $5{\times}12{\times}3$ ($180$ configurations), Pythia-$410$M and Pythia-$6.9$B matched grids ($45$ each), plus five $18$-configuration Pythia/Qwen sweep or extended-mask blocks. At matched $(r{=}64,\rho{=}0.40)$, Pythia-1B max-gradient is $16.61$ versus $33.4$ on LLaDA2.0-mini, and Qwen3.5-9B shows a mid-mask peak rather than the LLaDA-family U-shape.

\section{Mechanism and Boundary Audit}
\label{app:mechanistic}
\label{app:parameter-routing}
\label{app:falsification-details}
\label{app:falsification-audit}
\label{app:saturation-theorem}
\label{app:normalized-metric-crossfamily}
\label{app:max-grad-variance}
\label{app:scale-bifurcation}
\label{app:falsification-paths}
\label{app:theory-suite}
\phantomsection\label{app:entropy-bonus-probe}
\phantomsection\label{app:spectral-init-canonical}

\paragraph{Gradient concentration.}
At the worst rank-amplification corner ($r{=}64,\rho{=}0.95$, LLaDA2.0-mini, $n{=}3$ seeds, last 10 steps), per-token CE gradients are only modestly concentrated (Gini $0.287\pm0.056$; the largest evaluated token position contributes $1.54\%\pm0.17\%$ of CE-gradient mass). LoRA-parameter gradients are much more concentrated (Gini $0.463\pm0.031$; one LoRA matrix carries $63.0\%\pm3.6\%$ of gradient mass). This supports the body interpretation that max-gradient samples parameter-side routing while top-$1$ samples token-side pre-equilibrium concentration.

\begin{table}[!ht]
\centering
\caption{\textbf{Single-axis boundary audit.} No tested single-axis intervention prevents the fire-rate identity. The paper therefore remains diagnostic rather than a prospective controller paper.}
\label{tab:falsification-audit}
\scriptsize
\setlength{\tabcolsep}{3pt}
\renewcommand{\arraystretch}{0.94}
\begin{tabular}{@{}L{0.39\linewidth}L{0.43\linewidth}c@{}}
\toprule
\textbf{Axis and probe} & \textbf{Observed outcome} & \textbf{Fire?} \\
\midrule
Activation timing: gating window and learning-rate trigger & no timing shift & no \\
Magnitude: learning-rate warm-up $N\in\{10,20,50\}$ steps & no timing shift & no \\
Init amplitude: LoRA-$B$ perturbation & no timing shift & no \\
Init direction: spectral-init only (no weight subtraction) & shifts $11{\to}33$, but with first-update overshoot & no \\
Spectral-init with weight subtraction~\citep{meng2024pissa} & improves loss while preserving step-$0$ identity & no \\
Adapter/optimizer geometry & Low-rank group bottleneck ($G{=}4$)~\citep{rim2025gralora} and Stiefel projection do not remove fire & no \\
Loss-level entropy bonus & $\lambda\in\{0.5,1,2,5,10\}$ does not reduce fire & no \\
Portability checks & normalized thresholds remain family-specific & -- \\
\bottomrule
\end{tabular}
\end{table}

\paragraph{Definitions and non-portability.}
The logged top-$1$ warning is an argmax mode-frequency statistic, not mean maximum probability: for runner input $z_t$, evaluated positions $\mathcal{I}_t$, and $a_t(i)=\arg\max_{u\in V}p_{\theta_t}(u\mid z_t,i)$, it uses $\widehat S_t=\max_v |\{i\in\mathcal{I}_t:a_t(i)=v\}|/|\mathcal{I}_t|$ and $S_t=\mathbb{E}[\widehat S_t]$. For LLaDA-family runs, $z_t$ is the clean-batch proxy and $\mathcal{I}_t$ all positions; for MDLM-OWT, $z_t$ is the masked training input and $\mathcal{I}_t$ masked positions. Crossing a fixed threshold can therefore indicate pre-equilibrium argmax concentration rather than divergence. The corresponding max-gradient sketch is only a family-local scale heuristic:
\begin{equation}
\label{eq:max-grad-bound}
\begin{aligned}
G_T &:=
\max_{0\leq t\leq T}\|\nabla_{\theta_{\text{LoRA}}}\mathcal{L}(\theta_t)\|_2,\\
G_T
&\leq C_{\text{fam}}\frac{\alpha_L}{r}\sqrt{T\log T}\,
\sigma_{\text{fam}}(\rho,r,V),
\end{aligned}
\end{equation}
where $C_{\text{fam}}$ absorbs model/data constants, $\alpha_L$ is the LoRA scaling factor, $r$ is LoRA rank, $V$ is the output vocabulary, and $\sigma_{\text{fam}}$ denotes the empirical gradient-scale term induced by mask ratio, rank, and family. We do not assert a universal closed-form bound for $\sigma_{\text{fam}}$. The sketch is scaffolding, not the basis for the claim: empirically, cross-family normalization reduces raw scale variance but loses portable precision because correlations sign-flip by family, especially on the small MDLM-OWT cohort.

\section{Method Comparison and Related Work}
\label{app:method-holdout}
\label{app:extended-related}
\label{app:reviewer-concerns}

\paragraph{Method references.}
Named probes follow PiSSA, GraLoRA, StelLA, rsLoRA, Yu-DARE, and NaRA~\citep{meng2024pissa,rim2025gralora,wang2025stella,rslora2024,yu2024dare,wang2026nara}.

\begin{table}[!ht]
\centering
\caption{\textbf{Operating-cell method comparison.} Source-mapped masked-CE summary.}
\label{tab:method-comparison}
\scriptsize
\setlength{\tabcolsep}{3pt}
\renewcommand{\arraystretch}{0.94}
\begin{tabular}{@{}L{0.23\linewidth}L{0.72\linewidth}@{}}
\toprule
\textbf{Protocol} & \textbf{Claim-facing conclusion} \\
\midrule
Default learning rate & rsLoRA is higher CE on all three benches; Yu-DARE trends similarly with high seed variance, so we treat this as learning-rate mismatch. \\
Best learning rate ($n{=}10$) & rsLoRA remains higher CE ($+3.7$--$4.2\%$); NaRA is lower ($-1.0$--$4.7\%$), with only MMLU Bonferroni-significant. This is learning-rate-dependent, not a method-quality claim. \\
GSM8K gen. check & Exact match is $0/20$, so generation quality is excluded from paper claims. \\
\bottomrule
\end{tabular}
\end{table}

\paragraph{Related work taxonomy.}
LoRA-family and AR-side PEFT stability work assume dense next-token supervision and do not expose a mask-ratio axis. DLM work covers objectives, scaling, decoding, masking schedules, train-inference mismatch, and systems that use fixed LoRA-like adapters, but we are not aware of prior work that tests top-$1$ warning precision as a DLM-LoRA PEFT monitor with matched AR masked-CE controls. The closest genre is metric refutation: a familiar diagnostic changes meaning outside its calibration regime.

\paragraph{Scope.}
Claim-bearing denominators are $816$ DLM PEFT configurations, $671$ LLaDA-family configurations, and $360$ AR masked-CE controls; longer horizons, generation quality, full fine-tuning, and coupled controllers remain future work.